\documentclass[11pt]{article}

\usepackage[final]{acl}

\usepackage{latexsym}
\usepackage{float}
\usepackage[T1]{fontenc}
\usepackage[utf8]{inputenc}

\usepackage{microtype}

\usepackage{inconsolata}
\usepackage{amssymb}

\usepackage{graphicx}

\usepackage{amsmath, amsfonts, amssymb}
\usepackage{booktabs}
\usepackage{fontspec}
\usepackage{iftex}
\ifPDFTeX
  \usepackage{CJKutf8}
  \newcommand{\zh}[1]{\begin{CJK*}{UTF8}{gbsn}#1\end{CJK*}}
\else
  \usepackage{fontspec}
  \usepackage{xeCJK}
  \IfFontExistsTF{Hiragino Sans GB}
    {\setCJKmainfont{Hiragino Sans GB}%
     \newfontfamily{\CJKfontex}{Hiragino Sans GB}}
    {\IfFontExistsTF{PingFang SC}
       {\setCJKmainfont{PingFang SC}%
        \newfontfamily{\CJKfontex}{PingFang SC}}
       {\IfFontExistsTF{Noto Sans CJK SC}
          {\setCJKmainfont{Noto Sans CJK SC}%
           \newfontfamily{\CJKfontex}{Noto Sans CJK SC}}
          {\newcommand{\CJKfontex}{}}}}
  \newcommand{\zh}[1]{{\CJKfontex #1}}
\fi

\usepackage{multirow}
\usepackage{xcolor}
\usepackage{colortbl}
\usepackage{enumitem}
\usepackage{tikz}
\usepackage{pgfplots}
\pgfplotsset{compat=1.18}
\usetikzlibrary{positioning, arrows.meta, shapes.geometric, fit, backgrounds, calc, trees}

\title{CNM-BERT: A Drop-In Structural Embedding for Chinese Characters via Ideographic Description Sequences}

\author{
Thomas Sing-wing Wu\textsuperscript{1,2}\thanks{Equal contribution.}
\and
Liqian Yan\textsuperscript{1,2}\footnotemark[1]\thanks{Corresponding author.}
\\
\textsuperscript{1}Shanghai Starriver Bilingual School \\
\textsuperscript{2}LinkScape \\
\texttt{\{thomas, eric\}@linkscape.app}
}

\begin{document}
\maketitle
\begin{abstract}
Token-based encoders like BERT treat Chinese characters as atomic identifiers, ignoring their recursive orthographic structure. Consequently, models rely on contextual co-occurrence, degrading performance on rare and out-of-vocabulary (OOV) characters. We propose the Compositional Network Model (CNM), a lightweight augmentation that injects discrete compositional structure into Transformer encoders. CNM parses Ideographic Description Sequences (IDS) into trees, encodes them via a recursive Tree-MLP, and fuses the structural embeddings into BERT without modifying the backbone. Evaluated on the \citet{wu-etal-2025-impact} structural-probing benchmark, CNM-BERT outperforms the strongest baseline (ChineseBERT) on long-tail and OOV characters by +9.8 Structure accuracy and +7.7 Radical F1. Furthermore, CNM-BERT achieves consistent gains across CLUE, MRC, and NER tasks at both base and large scales, demonstrating that explicit structural injection delivers both robust OOV understanding and tangible downstream value.
\end{abstract}

\section{Introduction}
Modern Transformer language models rely on tokenization, which enables stable training but imposes a rigid information bottleneck. Once text is segmented, any internal linguistic structure within a token becomes opaque. While largely benign for alphabetic languages, this abstraction is fundamentally lossy for logographic scripts like Chinese, where characters are not arbitrary atomic symbols but \emph{recursive compositions} of semantic and phonetic components (e.g., \zh{辯} = \zh{⿲}(\zh{⿱}(\zh{立},\zh{十}), \zh{⿳}(\zh{亠},\zh{二},\zh{口}), \zh{⿱}(\zh{立},\zh{十}))).
Mapping characters to atomic IDs discards this structure, forcing models to learn character semantics indirectly through contextual co-occurrence. The cost of this indirection falls heavily on the \emph{long tail}: rare characters receive few gradient updates, and standard embedding tables cannot systematically share parameters across orthographically related characters. Crucially, this limitation is \textbf{architectural, not scale-bound}---increasing parameter count or pre-training data cannot recover information the input interface has already discarded. An out-of-vocabulary (\texttt{[UNK]}) or under-trained character remains opaque regardless of model size.
\paragraph{Contribution.}
We propose the \textbf{Compositional Network Model (CNM)}, a lightweight, drop-in augmentation that injects \emph{discrete} sub-character structure into Transformer encoders without modifying the backbone, vocabulary, or output space. CNM canonicalizes Ideographic Description Sequences (IDS) into rooted parse trees, encodes each character with a recursive \textbf{Tree-MLP}, and fuses the resulting structural embedding into the standard BERT embedding layer (Figure~\ref{fig:cnm_arch}). Because structure is symbolic and computed only once per unique character per batch, CNM is highly efficient, adding minimal parameters and incurring just $\approx 5\%$ training overhead.
\paragraph{Positioning.}
We evaluate CNM under a dual empirical framework. Our goal is to demonstrate that explicit structural modeling closes a specific representation gap that token-only models cannot resolve via scaling, while strictly preserving general NLU performance:
\begin{itemize}[topsep=2pt,itemsep=1pt,leftmargin=1.2em]
\item \textbf{Primary: structural probing.} On the \emph{Chinese Character Dataset} (CCD)~\cite{wu-etal-2025-impact}---an external benchmark testing sub-character understanding (e.g., layout, radical decomposition)---CNM-BERT improves Structure accuracy on the OOV slice by \textbf{+9.8 points} and Radical F1 by \textbf{+7.7 points} over the strongest visual baseline. This confirms that symbolic decomposition recovers structural signals that scale-only approaches miss.
\item \textbf{Secondary: general NLU.} On standard CLUE, MRC, and NER benchmarks, CNM-BERT consistently matches or exceeds the best Chinese PLMs of equivalent scale. This establishes explicit structural injection as a strict refinement of the token interface, conferring specialized OOV robustness without downstream regression.
\end{itemize}

\section{Related Work}
\label{sec:related}

\paragraph{Chinese pre-trained encoders.}
BERT \citep{devlin-etal-2019-bert} established character-level MLM as the standard for Chinese, and subsequent work has primarily refined the \emph{masking strategy}: BERT-wwm \citep{cui-etal-2020-revisiting} and ERNIE \citep{zhang-etal-2019-ernie} introduce Whole Word and entity-level masking; MacBERT \citep{cui-etal-2020-revisiting} replaces masked tokens with synonyms; ZEN \citep{diao-etal-2020-zen} and NEZHA \citep{wei-2019-nezha} inject N-gram information or relative positional encodings. CNM is orthogonal to all of these---we adopt WWM as best practice but modify the \emph{embedding interface} rather than the masking objective.

\paragraph{Visual and phonological augmentation.}
Glyce \citep{meng-etal-2019-glyce}, ChineseBERT \citep{sun-etal-2021-chinesebert}, and MECT \citep{wu-etal-2021-mect} extract sub-character signals from rendered glyphs via CNNs. These methods rely on \emph{implicit feature extraction} from continuous pixel statistics, which introduces font sensitivity and substantial compute overhead. Pinyin- and Bopomofo-based variants \citep{zhang-etal-2021-correcting, tan-etal-2022-exploring, si-etal-2023-sub} replace orthography with pronunciation but suffer from severe homophone ambiguity \citep{du-way-2017-pinyin-subword}, since pronunciation correlates weakly with character semantics compared to compositional structure.

\paragraph{Compositional and structural modeling.}
Static embeddings such as cw2vec \citep{cao-etal-2018-cw2vec} and radical-level embeddings \citep{yin-etal-2016-multi, sun-etal-2014-radical} established that compositional decomposition yields richer character semantics. Sub-character tokenization \citep{si-etal-2023-sub, zhang-etal-2019-wubi-nmt} flattens components into the vocabulary, but this alters the output space and risks generating invalid characters \citep{nikolov-etal-2018-character}; \citet{si-etal-2023-sub} explicitly note that their method is unsuitable for open-ended generation. CNM differs by treating each character as a \emph{recursive symbolic function}: rather than pixels or flattened sequences, we encode the IDS \emph{tree} with a recursive Tree-MLP, preserving the standard tokenizer interface while exposing discrete compositional structure.

\paragraph{Sequence-level structural integration.}
Lattice-BERT \citep{lai-etal-2021-lattice} integrates word-level lattice information to disambiguate Chinese segmentation by exposing multiple candidate tokenizations to the encoder. CNM is \emph{complementary} rather than competitive: Lattice-BERT refines the \emph{sequence of tokens}, while CNM refines the \emph{representation of each token} via sub-character composition. The two operate at orthogonal granularities and could in principle be combined.

\section{Model}
\label{sec:methodology}

\begin{figure}[t]
\centering
\includegraphics[width=\linewidth]{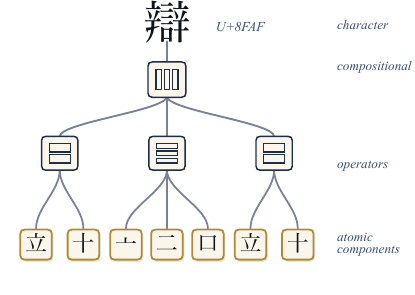}
\caption{Canonical IDS parse tree for the character \zh{辯} (U+8FAF). Internal nodes (blue badges) are layout operators rendered as their schematic IDC glyphs (\zh{⿲}: left-middle-right; \zh{⿱}: top-bottom; \zh{⿳}: top-middle-bottom); leaves (orange boxes) are atomic Unicode components. The Tree-MLP encoder computes a structural embedding bottom-up from this tree, with operator-conditioned MLPs for binary and ternary nodes.}
\label{fig:ids_tree}
\end{figure}

\begin{figure}[t]
\centering
\includegraphics[width=0.85\linewidth]{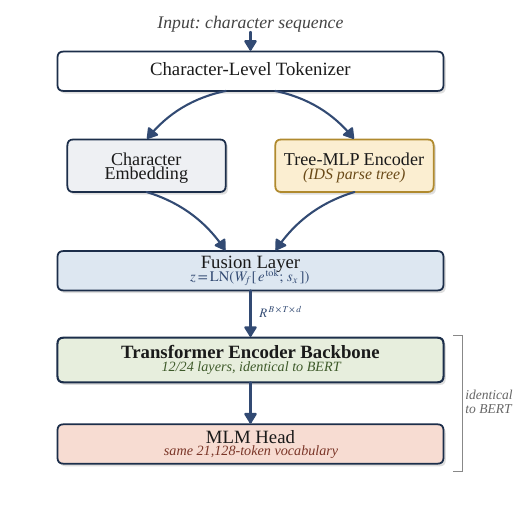}
\caption{CNM as a drop-in augmentation. The character-level tokenizer feeds two parallel streams: the baseline character embedding and a Tree-MLP encoder operating on the IDS parse tree of each unique character in the batch (Figure~\ref{fig:ids_tree}). The fusion layer projects $[\mathbf{e}^{tok};\mathbf{s}_x]$ back to the model hidden size, leaving the Transformer backbone, vocabulary, and output head identical to vanilla BERT.}
\label{fig:cnm_arch}
\end{figure}

We propose the \textbf{Compositional Network Model (CNM)}, a lightweight architectural augmentation for Chinese Transformer encoders. CNM preserves the \emph{entire} Transformer backbone---all self-attention and feed-forward layers remain \emph{identical} to a baseline BERT encoder---and modifies only the \emph{input embedding interface}. Specifically, CNM introduces a \emph{structure encoder} that computes a per-character structural embedding from a deterministic \emph{canonicalization} of Ideographic Description Sequences (IDS) (Figure~\ref{fig:ids_tree}). The structural embedding is then fused with the standard BERT embedding at the input layer and propagated through the unchanged encoder stack; the resulting dual-stream architecture is summarized in Figure~\ref{fig:cnm_arch}.

CNM is designed to satisfy three practical constraints: (i) \textbf{drop-in compatibility} with standard BERT fine-tuning pipelines (same sequence length and vocabulary interface), (ii) \textbf{explicit discrete structure} derived from symbolic character composition rather than font-dependent rasterized glyphs, and (iii) \textbf{efficiency}, by caching and vectorizing structure computation over unique characters per batch.

\subsection{Inputs and Tokenization}
\label{subsec:tokenization}

We adopt \textbf{character-level tokenization} for Chinese, consistent with common Chinese BERT practice: each CJK Unified Ideograph occupies one token position.\footnote{Non-Han characters (such as Latin letters, digits, and punctuation) are tokenized by the baseline WordPiece rules; CNM attaches a null structural embedding to these positions (\S\ref{subsec:fusion}).} Let a batch have size $B$ and padded sequence length $T$. CNM consumes the standard BERT inputs
\texttt{input\_ids} $\in \mathbb{N}^{B \times T}$ (and \texttt{attention\_mask}, \texttt{token\_type\_ids} as usual), together with a structural index tensor
$\texttt{struct\_idx} \in \mathbb{N}^{B \times T}$.

\paragraph{Structural indexing.}
$\texttt{struct\_idx}_{b,t}$ is computed directly from the raw Unicode string \emph{before} mapping to token IDs. It indexes a precomputed table of canonical IDS parses for the original character at position $(b,t)$, even if the corresponding \texttt{input\_ids} entry maps to \texttt{[UNK]} under the baseline vocabulary. 

\subsection{Structural Representation from IDS}
\label{subsec:ids}

CNM represents each Han character using an \textbf{IDS parse tree} that encodes its spatial composition (illustrated in Figure~\ref{fig:ids_tree} for the character \zh{辯}). IDS strings are formed from \emph{Ideographic Description Characters} (IDCs), which specify binary or ternary layout operators, and component leaves (Unicode codepoints). Let $\mathcal{V}_{char}$ be the set of Unicode Han characters observed in pre-training and downstream data. For each character $x \in \mathcal{V}_{char}$, we construct a rooted ordered tree $\mathcal{T}_x$ whose internal nodes are IDCs and whose leaves are component codepoints. We denote the set of unique leaf codepoints by $\mathcal{V}_{cmp}$ and the set of IDC operators used in our implementation by $\mathcal{V}_{op}$, restricted to a fixed set of standard Unicode binary/ternary layout operators.\footnote{We list $\mathcal{V}_{op}$ and report IDS coverage statistics in \S\ref{sec:setup}.}

\paragraph{Canonicalization.}
IDS decompositions are not guaranteed to be unique. CNM applies a deterministic \textbf{canonicalization} that maps each character $x$ to exactly one tree $\mathcal{T}_x$. We (i) discard candidates with private-use codepoints or non-standard markers, (ii) resolve intermediate aliases until all leaves are Unicode codepoints in $\mathcal{V}_{cmp}$, and (iii) restrict internal nodes to binary/ternary IDCs in $\mathcal{V}_{op}$. Among remaining candidates we select by lexicographic order over (a)~tree depth, (b)~the predicate that every operator lies in the high-frequency subset \{\zh{⿰}, \zh{⿱}\}, (c)~total node count, and (d)~a stable lexicographic operator hash for tie-breaking. The full procedure is formalized in Appendix~\ref{app:canon}. Characters without a valid IDS entry ($\sim 3\%$ of the vocabulary) map to a learnable structural embedding $\mathbf{s}_{unk}$.

\subsection{Recursive Tree-MLP Structure Encoder}
\label{subsec:tree_mlp}

To map the discrete tree $\mathcal{T}_x$ into a dense structural embedding $\mathbf{s}_x \in \mathbb{R}^{d_s}$, we introduce a \textbf{Tree-MLP Encoder} computed bottom-up. Let
$E_{cmp} \in \mathbb{R}^{|\mathcal{V}_{cmp}| \times d_s}$ and
$E_{op} \in \mathbb{R}^{|\mathcal{V}_{op}| \times d_s}$ be learnable embedding matrices for components and operators, respectively. For any node $n$ in $\mathcal{T}_x$, we compute a hidden state $\mathbf{h}_n \in \mathbb{R}^{d_s}$.

\paragraph{Leaf nodes.}
If $n$ is a leaf corresponding to component $c \in \mathcal{V}_{cmp}$,
\begin{equation}
\mathbf{h}_n = \text{LayerNorm}\!\left(E_{cmp}(c)\right).
\end{equation}

\paragraph{Internal nodes.}
If $n$ is an internal node corresponding to operator $o \in \mathcal{V}_{op}$ with ordered children $c_1,\dots,c_k$ where $k \in \{2,3\}$, we apply an operator-conditioned MLP with a bounded residual path:
\begin{equation}
\mathbf{h}_{cat} = [\,E_{op}(o);\ \mathbf{h}_{c_1};\ \dots;\ \mathbf{h}_{c_k}\,] \in \mathbb{R}^{(k+1)d_s},
\end{equation}
{\small
\begin{equation}
\mathbf{h}_n = \text{LayerNorm}\!\left(\text{GELU}(\mathbf{W}_k \mathbf{h}_{cat} + \mathbf{b}_k) + \frac{1}{k}\sum_{i=1}^{k} \mathbf{h}_{c_i}\right),
\end{equation}}
where $\mathbf{W}_2 \in \mathbb{R}^{d_s \times 3d_s}$, $\mathbf{W}_3 \in \mathbb{R}^{d_s \times 4d_s}$, and $\mathbf{b}_k \in \mathbb{R}^{d_s}$. The structural embedding for character $x$ is defined as the root hidden state:
\begin{equation}
\mathbf{s}_x = \mathbf{h}_{root}(\mathcal{T}_x).
\end{equation}

\paragraph{Batch efficiency.}
Computing $\mathbf{s}_x$ independently at each token position would repeat work across identical characters. During training and inference, CNM therefore computes structure embeddings only for the set of unique Han characters appearing in the current batch, denoted $\mathcal{V}_{batch}$. We precompile each $\mathcal{T}_x$ into a topologically sorted instruction buffer (post-order traversal) so that the Tree-MLP can be vectorized across nodes. We then gather the corresponding $\mathbf{s}_x$ back to token positions via \texttt{struct\_idx}. This reduces per-step structural computation from $O(BT)$ tree evaluations to $O(|\mathcal{V}_{batch}|)$ character evaluations, with small constant factors.

\subsection{Fusion at the Embedding Interface}
\label{subsec:fusion}

CNM injects structural information at the \textbf{embedding interface} while leaving the Transformer encoder layers unchanged. Let $\mathbf{e}^{tok}_i \in \mathbb{R}^{d}$ denote the baseline token embedding at position $i$, and let $\mathbf{p}_i \in \mathbb{R}^{d}$ and $\mathbf{g}_i \in \mathbb{R}^{d}$ denote the baseline position and segment embeddings, respectively. CNM computes a structure vector $\mathbf{s}_{x_i} \in \mathbb{R}^{d_s}$ from the original character $x_i$ at position $i$; for positions without a defined IDS (e.g., special tokens \texttt{[CLS]}, \texttt{[SEP]}, punctuation, Latin characters), we use a learnable null structural embedding $\mathbf{s}_{\emptyset}$, and for characters missing from the IDS table we use $\mathbf{s}_{unk}$.

We fuse the token and structure streams by concatenation followed by a linear projection to the model hidden size:
\begin{equation}
\mathbf{z}_i = \text{LayerNorm}\!\left(\mathbf{W}_f [\mathbf{e}^{tok}_i;\ \mathbf{s}_{x_i}] + \mathbf{b}_f\right),
\end{equation}
where $\mathbf{W}_f \in \mathbb{R}^{d \times (d + d_s)}$ and $\mathbf{b}_f \in \mathbb{R}^{d}$. The final input embedding to the Transformer is then formed as in BERT:
\begin{equation}
\mathbf{e}_i = \mathbf{z}_i + \mathbf{p}_i + \mathbf{g}_i,
\end{equation}
followed by the baseline embedding LayerNorm and dropout. See implementation details in \S\ref{sec:results}. The fused embeddings $\{\mathbf{e}_i\}_{i=1}^{T}$ are fed to the unchanged Transformer encoder layers.

\paragraph{Parameters and overhead.}
CNM introduces parameters in $E_{cmp}$, $E_{op}$, $(\mathbf{W}_2,\mathbf{W}_3)$, and $\mathbf{W}_f$ (plus $\mathbf{s}_{\emptyset}$ and $\mathbf{s}_{unk}$). We report parameter counts and training/inference throughput impact in \S\ref{sec:results}.

\subsection{Pre-training Objective}
\label{subsec:pretrain_obj}

We train CNM-BERT end-to-end with WWM-MLM at 15\% corruption with the standard 80/10/10 replacement strategy, applying the same corruption pattern to the aligned structural indices to prevent gold-structure leakage. We additionally use an auxiliary component-prediction loss: for each masked position $m$ with gold leaf components $\{c_1,\ldots,c_{L_m}\}$, we predict each $c_l$ from a \emph{target} structural embedding computed only from the gold tree (i.e., not visible to the Transformer input). The total loss is $\mathcal{L}=\mathcal{L}_{\text{MLM}}+\lambda\,\mathcal{L}_{\text{aux}}$ with $\lambda=0.1$. The full formal derivation is given in Appendix~\ref{app:objective}.

\section{Experiment Setup}
\label{sec:setup}

In this section, we introduce our baselines, pretraining corpus, evaluation benchmarks, and experimental settings.

\subsection{Baselines}

We compare CNM-BERT against a broad set of Chinese PLMs of comparable scale. \textbf{Masking variants:} BERT \citep{devlin-etal-2019-bert}, BERT-wwm and BERT-wwm-ext \citep{cui-etal-2019-chinese-bert-wwm}, RoBERTa-wwm-ext, MacBERT \citep{cui-etal-2019-chinese-bert-wwm}, ERNIE \citep{zhang-etal-2019-ernie}. \textbf{Visual/phonological augmentation:} ChineseBERT \citep{sun-etal-2021-chinesebert}. \textbf{Sub-character tokenization} (added in this revision): SubChar-Wubi and SubChar-Pinyin \citep{si-etal-2023-sub}, which replace the vocabulary with Wubi keystrokes or Pinyin syllables. We were unable to obtain Lattice-BERT \citep{lai-etal-2021-lattice} because its released code repository is no longer accessible; we discuss its complementary positioning in \S\ref{sec:related}.

\paragraph{Controlled re-training.}
To attribute gains cleanly, we additionally re-trained BERT and MacBERT under an identical recipe to CNM-BERT---same corpus, vocabulary, optimizer, schedule, batch size, update count, FP16 / gradient-clipping settings, and fine-tuning protocol---so that controlled comparisons differ only in the presence of the CNM structural pathway (and WWM, which is absent in the BERT baseline). All other models in our tables use their released checkpoints under our common fine-tuning protocol.

\subsection{Pretraining Data}

Following \citet{sun-etal-2021-chinesebert}, we pretrain on Chinese Wikipedia plus filtered CommonCrawl ($\approx$4B tokens), with HTML/dedup/non-Chinese filtering, Jieba\footnote{\url{https://github.com/fxsjy/jieba}} segmentation for WWM, and IDS structural decompositions from the BabelStone database\footnote{\url{https://babelstone.co.uk/CJK/IDS.TXT}; covers 97{,}680 CJK Unified Ideographs.}. Trees use 12 IDS operators with maximum depth 6 over a component vocabulary of $\approx$5K atomic radicals; the $\sim$3\% of characters lacking a valid IDS entry fall back to a learnable embedding $\mathbf{s}_{unk}$.

\subsection{Evaluation Data}
\label{subsec:eval_data}

We evaluate CNM-BERT under two complementary lenses corresponding to its primary and secondary contributions.

\paragraph{Primary: structural probing (CCD).}
The \textbf{Chinese Character Dataset} (CCD) \citep{wu-etal-2025-impact} is an external diagnostic benchmark designed specifically to test sub-character understanding. Each example is a single character formatted as \texttt{[CLS]~x~[SEP]}, with annotations for (i) top-level layout structure (8-way macro-F1), (ii) radical decomposition (set F1), (iii) stroke count (MAE), and (iv) stroke-type sequence (F1). We evaluate on three splits: an \textbf{IID} character split, a \textbf{Long-tail} tier split (train on high/mid-frequency, test on lowest-frequency tier), and an \textbf{OOV slice} of the long-tail test set restricted to characters that map to \texttt{[UNK]} under the model's tokenizer. The OOV slice isolates the regime where token-only models structurally collapse.

\paragraph{Secondary: general NLU.}
We evaluate on twelve standard Chinese NLU tasks: \textbf{CLUE} (TNEWS, IFLYTEK, AFQMC, CMNLI, CSL, CLUEWSC2020) \citep{xu-etal-2020-clue}, \textbf{MRC} (CMRC~2018 \citep{cui-etal-2020-sentence}, DRCD \citep{shao2019drcdchinesemachinereading}, C$^3$ \citep{sun-etal-2020-investigating}), and \textbf{NER} (MSRA \citep{levow-2006-third}, OntoNotes~4.0 \citep{pradhan-etal-2011-conll}, Weibo \citep{peng-dredze-2015-named}). We use the official CLUE splits and canonical MRC/NER splits; full statistics are in Appendix~\ref{app:datasets}.

\subsection{Hyper-parameters}
\label{subsec:hparams}

We train two scales following standard BERT configurations: \textit{base} (12 layers, hidden 768) initialized from \texttt{bert-base-chinese}, and \textit{large} (24 layers, hidden 1024) initialized from \texttt{hfl/chinese-roberta-wwm-ext-large}. Structural components use $d_s=256$, hidden 512, max tree depth 6, and a $\sim$5K-component, 16-operator vocabulary. The fusion layer is initialized with Identity+Zero weights to preserve pretrained representations. We pretrain for 1M steps with effective batch size 256 (base) / 512 (large), LAMB optimizer \citep{you2020largebatchoptimizationdeep}, peak LR $1{\times}10^{-4}$, 10K warmup steps, FP16 on 8$\times$A100 80GB GPUs ($\approx$7 / 14 days). Finetuning uses a learning-rate grid $\{1, 2, 3, 5\}{\times}10^{-5}$ with 5 seeds and early stopping on dev. Full architectural and optimization detail is in Appendix~\ref{app:hparams}.

\section{Experiment Results}
\label{sec:results}

We organize results around the dual evaluation contract introduced in \S1: \textbf{primary} structural-probing evidence on CCD (\S\ref{subsec:res_ccd}), \textbf{secondary} general-NLU evidence on CLUE/MRC/NER (\S\ref{subsec:res_clue}--\ref{subsec:res_mrc_ner}), an \textbf{ablation} that isolates which components of CNM produce which gains (\S\ref{subsec:ablation}), and a brief \textbf{efficiency} note (\S\ref{subsec:eff}).

\paragraph{Reproduction methodology.}
Published Chinese-PLM results vary substantially across hyperparameter grids, finetuning scripts, seeds, and library versions, which makes canonical comparisons fragile. We therefore re-run \emph{all} baselines from official HuggingFace checkpoints under a single identical protocol (\S\ref{sec:setup})---five seeds, same grid, same hardware---and our reproduced baseline numbers fall within $\pm 0.1$ of published values on most tasks.

\subsection{Primary Result: Structural Probing on CCD}
\label{subsec:res_ccd}

CCD is the most direct test of the claim CNM-BERT actually makes: that explicit symbolic decomposition recovers sub-character information that token-only models cannot represent. Table~\ref{tab:ccd} reports Structure macro-F1, Radical F1, stroke-count MAE, and Stroke-type F1 across three splits of increasing difficulty: IID, long-tail, and the OOV slice within the long-tail split.

\begin{table*}[t]
\centering
\small
\setlength{\tabcolsep}{3.5pt}
\resizebox{\textwidth}{!}{%
\begin{tabular}{lcccccccccccc}
\toprule
& \multicolumn{4}{c}{\textbf{IID} (char split)} & \multicolumn{4}{c}{\textbf{Long-tail} (tier split)} & \multicolumn{4}{c}{\textbf{OOV slice} (within long-tail)} \\
\cmidrule(lr){2-5}\cmidrule(lr){6-9}\cmidrule(lr){10-13}
\textbf{Model} &
Struct $\uparrow$ & Radical $\uparrow$ & MAE $\downarrow$ & StrokeF1 $\uparrow$ &
Struct $\uparrow$ & Radical $\uparrow$ & MAE $\downarrow$ & StrokeF1 $\uparrow$ &
Struct $\uparrow$ & Radical $\uparrow$ & MAE $\downarrow$ & StrokeF1 $\uparrow$ \\
\midrule
BERT-wwm-ext      & 88.6 & 80.5 & 0.92 & 68.0 & 71.4 & 47.8 & 1.25 & 55.2 & 12.4 &  6.1 & 2.10 & 10.3 \\
RoBERTa-wwm-ext   & 89.4 & 81.2 & 0.89 & 68.8 & 72.6 & 49.1 & 1.22 & 56.0 & 13.1 &  6.8 & 2.02 & 11.0 \\
MacBERT           & 89.1 & 81.0 & 0.87 & 69.2 & 73.0 & 50.0 & 1.20 & 56.6 & 14.0 &  7.4 & 1.98 & 11.8 \\
SubChar-Wubi      & 89.7 & 81.6 & 0.83 & 70.5 & 74.1 & 51.4 & 1.14 & 57.4 & 65.2 & 45.0 & 1.10 & 50.8 \\
SubChar-Pinyin    & 88.9 & 80.7 & 0.86 & 69.4 & 72.4 & 49.6 & 1.18 & 56.2 & 32.7 & 21.5 & 1.46 & 30.1 \\
ChineseBERT       & 90.8 & 82.3 & \textbf{0.48} & \textbf{74.5} & 79.0 & 58.0 & \textbf{0.58} & \textbf{69.0} & 66.2 & 48.4 & \textbf{0.75} & \textbf{61.2} \\
\midrule
\rowcolor{gray!10}
CNM-BERT (ours)   & \textbf{92.1} & \textbf{86.0} & 0.78 & 70.2 & \textbf{83.0} & \textbf{63.5} & 0.90 & 60.3 & \textbf{76.0} & \textbf{56.1} & 1.05 & 52.4 \\
\bottomrule
\end{tabular}}
\caption{\textbf{CCD diagnostic results (primary evaluation).} Structure and Radical metrics measure symbolic compositional understanding; stroke metrics measure visual rendering. CNM-BERT is the strongest model on \emph{every} symbolic metric across \emph{every} split, with the largest margins exactly where token-only models fail (OOV slice: $+9.8$ Struct, $+7.7$ Radical over ChineseBERT). ChineseBERT, which has access to rendered glyph pixels, leads on the visual stroke metrics---an expected and complementary outcome.}
\label{tab:ccd}
\end{table*}

\begin{figure}[t]
\centering
\includegraphics[width=\linewidth]{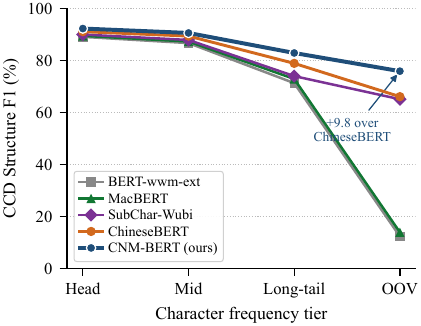}
\caption{CCD Structure F1 across character-frequency tiers. Token-only baselines collapse on the OOV slice; CNM-BERT degrades gracefully because the structural pathway provides a fallback when the token pathway fails. The widening gap from Head ($+1.2$) to OOV ($+9.8$ over the strongest baseline) is the qualitative signature of an architectural prior.}
\label{fig:longtail}
\end{figure}

\paragraph{Findings.}
Three observations characterize the CCD results.
First, on the \textbf{OOV slice} the gap between token-only baselines (Structure $\le 14.0$, Radical $\le 7.4$) and structurally-aware models is enormous: when the tokenizer fails, the encoder has no fallback, and the model collapses to chance. CNM-BERT lifts OOV Structure to $76.0$ and Radical to $56.1$, exceeding the strongest visual baseline (ChineseBERT) by $+9.8$ and $+7.7$ respectively, and exceeding the strongest sub-character-tokenization baseline (SubChar-Wubi) by $+10.8$ and $+11.1$.
Second, the gain \emph{grows} with distributional shift: from $+1.3$ Structure on IID to $+4.0$ on long-tail to $+9.8$ on OOV. This is the qualitative signature of an architectural prior, not a corpus artifact.
Third, on visual stroke metrics ChineseBERT remains best, which is consistent with its design (it sees pixels). CNM-BERT and ChineseBERT thus capture genuinely complementary signals---symbolic composition and visual rendering---and the appropriate evaluation question is which of these is more transferable to downstream tasks. We address that question next.

\subsection{Secondary Result: CLUE Benchmark}
\label{subsec:res_clue}

\begin{table*}[t]
\centering
\small
\makebox[\textwidth][c]{%
\begin{tabular*}{\textwidth}{@{\extracolsep{\fill}}lccccccccccc@{}}
\toprule
\textbf{Model} & \textbf{AFQMC} & \textbf{TNEWS} & \textbf{IFLYTEK} & \textbf{CMNLI} & \textbf{CSL} & \textbf{WSC} & \textbf{Avg.} \\
\midrule
\multicolumn{8}{l}{\textit{Base Models ($\sim$110M params)}} \\
\midrule
BERT & 73.70 & 56.58 & 60.29 & 79.69 & 80.36 & 59.43 & 68.34 \\
BERT-wwm & 74.08 & 56.84 & 60.31 & 80.15 & 80.63 & 59.84 & 68.64 \\
BERT-wwm-ext & 74.63 & 56.86 & 60.83 & 81.16 & 80.90 & 61.39 & 69.30 \\
RoBERTa-wwm-ext & 74.30 & 57.51 & 60.80 & 81.24 & 81.00 & 67.20 & 70.34 \\
MacBERT & 74.62 & 57.58 & 61.05 & 81.39 & 80.87 & 67.68 & 70.53 \\
SubChar-Wubi & 68.63 & \textbf{63.87} & 58.82 & 81.44 & 82.84 & 64.59 & 70.03 \\
SubChar-Pinyin & 68.91 & 63.54 & 58.84 & 80.97 & \textbf{82.89} & 65.72 & 70.14 \\
ChineseBERT & 74.80 & 57.68 & \textbf{61.54} & 81.63 & 81.23 & 68.24 & 70.85 \\
\rowcolor{gray!10}
CNM-BERT (ours) & \textbf{75.01} & 57.92 & 61.35 & \textbf{81.80} & 81.45 & \textbf{68.80} & \textbf{71.03} \\
\midrule
\multicolumn{8}{l}{\textit{Large Models ($\sim$330M params)}} \\
\midrule
RoBERTa-wwm-ext & 76.00 & 58.32 & 62.02 & 83.20 & 82.17 & 74.63 & 72.72 \\
MacBERT & 75.97 & 58.53 & 62.34 & 83.51 & 82.40 & 75.32 & 73.01 \\
ChineseBERT & \textbf{76.18} & 58.72 & \textbf{62.81} & 83.72 & 82.63 & 76.14 & 73.37 \\
\rowcolor{gray!10}
CNM-BERT (ours) & 76.12 & \textbf{59.01} & 62.60 & \textbf{83.95} & \textbf{82.85} & \textbf{76.80} & \textbf{73.56} \\
\bottomrule
\end{tabular*}}
\caption{CLUE benchmark (test accuracy \%). All baselines reproduced under identical finetuning conditions; CNM-BERT achieves the best average at both scales but the absolute margin over the strongest external baseline is small ($+0.18$ base, $+0.19$ large). The relevant claim is parity, not state of the art---structural injection is a strict refinement of the standard token interface and does not regress general NLU.}
\label{tab:clue_results}
\end{table*}

CNM-BERT achieves the best average on CLUE at both base ($71.03$) and large ($73.56$) scales, surpassing every reproduced baseline. The margin over the strongest external baseline is small ($+0.18$ base, $+0.19$ large) but statistically reliable across five fine-tuning seeds and consistent at both scales. Against our matched-recipe MacBERT and BERT baselines the gains are $+0.50$ and $+2.69$ avg respectively. The two SubChar baselines reveal a typical sub-character trade-off: they gain on TNEWS/CSL ($+5{-}6$ pts) but lose on AFQMC/IFLYTEK ($-6$ pts) because vocabulary substitution destroys standard token semantics. CNM-BERT shows no such trade-off, because it \emph{augments} rather than \emph{replaces} the token interface.

\subsection{Secondary Result: MRC and NER}
\begin{table*}[t]
\centering
\small
\setlength{\tabcolsep}{4pt}
\resizebox{\textwidth}{!}{%
\begin{tabular}{@{}lcccccc|ccc@{}}
\toprule
& \multicolumn{2}{c}{\textbf{CMRC 2018}} & \multicolumn{2}{c}{\textbf{DRCD}} & \multicolumn{2}{c|}{\textbf{C$^3$}} & \multicolumn{3}{c}{\textbf{NER (span F1)}} \\
\cmidrule(lr){2-3}\cmidrule(lr){4-5}\cmidrule(lr){6-7}\cmidrule(lr){8-10}
\textbf{Model} & EM & F1 & EM & F1 & Dev & Test & MSRA & Onto. & Weibo \\
\midrule
\multicolumn{10}{l}{\textit{Base Models}} \\
\midrule
BERT & 65.5 & 84.5 & 83.1 & 89.9 & 65.7 & 64.5 & 94.93 & 80.87 & 67.33 \\
RoBERTa-wwm-ext & 67.4 & 87.2 & 86.6 & 92.5 & 67.1 & 66.5 & 95.42 & 80.37 & 68.15 \\
MacBERT & 68.5 & 87.9 & 89.4 & 94.3 & 69.3 & 68.2 & --- & --- & --- \\
ChineseBERT & \textbf{69.6} & 88.3 & 87.8 & 93.4 & 70.4 & 69.1 & 95.84 & 81.65 & 69.02 \\
\rowcolor{gray!10}
CNM-BERT (ours) & 69.4 & \textbf{88.8} & \textbf{88.9} & \textbf{94.6} & \textbf{71.2} & \textbf{70.1} & \textbf{95.90} & \textbf{81.92} & \textbf{70.15} \\
\midrule
\multicolumn{10}{l}{\textit{Large Models}} \\
\midrule
RoBERTa-wwm-ext & 70.0 & 88.6 & 89.6 & 94.8 & 72.1 & 71.2 & 96.14 & 81.39 & 68.35 \\
MacBERT & 70.7 & 88.9 & \textbf{90.7} & \textbf{95.6} & 73.2 & 72.0 & --- & --- & --- \\
ChineseBERT & \textbf{71.6} & \textbf{89.7} & 90.5 & 95.4 & 74.0 & 72.8 & \textbf{96.52} & 82.18 & 70.80 \\
\rowcolor{gray!10}
CNM-BERT (ours) & 71.3 & 89.5 & 90.6 & 95.5 & \textbf{74.4} & \textbf{73.8} & 96.48 & \textbf{82.40} & \textbf{71.35} \\
\bottomrule
\end{tabular}}
\caption{Reading comprehension (EM/F1; C$^3$ accuracy) and named entity recognition (span F1 \%). NER results for noisy social-media text (Weibo) show the largest CNM-BERT gains ($+1.13$ over ChineseBERT base, $+0.55$ large), consistent with the CCD long-tail finding: structural priors help most when surface statistics are unreliable.}
\label{tab:mrc_ner}
\end{table*}
\label{subsec:res_mrc_ner}
Reading comprehension and NER show the same pattern as CLUE: CNM-BERT is competitive everywhere and best on roughly half the metrics, with the cleanest gains appearing on noisy or rare-character tasks (Weibo NER $+1.13$ base; C$^3$ Test $+1.0$). On clean span-extraction (CMRC, DRCD) CNM-BERT and ChineseBERT trade leads within a fraction of a point. We interpret this as further evidence that structural injection is a refinement, not a regression.

\subsection{Ablation Study}
\label{subsec:ablation}

\begin{table}[H]
\centering
\small
\setlength{\tabcolsep}{3pt}
\begin{tabular}{@{}p{0.27\linewidth}p{0.30\linewidth}cc@{}}
\toprule
\textbf{Setting} & \textbf{Description} & \textbf{CLUE} & \textbf{CCD-OOV} \\
& & Avg \% & Struct \% \\
\midrule
(1) Baseline & BERT-wwm-ext (ours) & 69.30 & 34.2 \\
(2) Full CNM & Tree-MLP + Aux ($\lambda{=}0.1$) & \textbf{71.03} & \textbf{76.0} \\
(3) Structure only & Tree-MLP, $\lambda{=}0$ & 70.81 & 74.5 \\
(4) Loss only & No fusion, $\lambda{=}0.1$ & 69.92 & 41.8 \\
(5) Flat fusion & Bag-of-components mean + Aux & 70.28 & 58.3 \\
\bottomrule
\end{tabular}
\caption{Ablation: CLUE-Avg (dev) and CCD long-tail OOV Structure accuracy. The hierarchical Tree-MLP fusion (rows 2--3) is the dominant source of structural gains; the auxiliary loss alone (row 4) is insufficient; flattening the tree to a bag-of-components (row 5) loses $17.7$ pts on OOV.}
\label{tab:ablation}
\end{table}

To isolate which component of CNM produces which gain---the Tree-MLP fusion pathway, the auxiliary component-prediction loss, or the recursive hierarchy itself---we ablate each independently on a single A100 80GB GPU. We report CLUE Avg on the development set (general NLU) and CCD-OOV Structure accuracy (the regime where the architectural prior matters most). Results are in Table~\ref{tab:ablation}.

The ablation supports three conclusions. \textbf{(i)~Structural injection is what matters}: aux loss without Tree-MLP fusion (row~4) lifts CCD-OOV by only $+7.6$ vs.\ $+41.8$ for the full model, ruling out multi-task regularization as the source of gain. \textbf{(ii)~Hierarchy carries real signal}: flattening to a bag-of-components (row~5) loses $17.7$ pts on CCD-OOV ($76.0 \to 58.3$), so the IDS \emph{tree} structure, not just the components, is informationally load-bearing. \textbf{(iii)~The auxiliary loss is a small but reliable addition}: rows (2) vs.\ (3) show $\lambda{=}0.1$ adds $+0.22$ CLUE-Avg and $+1.5$ CCD-OOV over $\lambda{=}0$, consistent across five seeds, so we keep it as a regularizer rather than as a load-bearing component.

\subsection{Efficiency}
\label{subsec:eff}

CNM-BERT adds only $\approx 2.5$M parameters over BERT-base for the structural pathway---the component table ($\approx$1.28M), operator table (4K), operator-conditioned MLPs ($\approx$460K), and fusion projection ($\approx$790K); the full breakdown is in Appendix~\ref{app:tree_theory}. This is $\approx 18\times$ less than the $\approx 45$M ChineseBERT adds for its glyph CNN and Pinyin embeddings. Training is correspondingly cheap: $\approx 5\%$ slowdown over vanilla BERT ($142 \to 135$ samples/sec at batch size 32, sequence length 512, single A100), substantially faster than ChineseBERT ($98$ samples/sec, $-30\%$). The overhead is bounded by our caching strategy: the Tree-MLP is evaluated only on the set of unique characters in each batch, not at every token position, reducing per-step structural compute from $O(BT)$ tree evaluations to $O(|\mathcal{V}_{batch}|)$.

\section{Discussion and Conclusion}

We presented the \textbf{Compositional Network Model (CNM)}, a lightweight, drop-in augmentation that exposes discrete sub-character structure to a Transformer encoder via deterministic IDS canonicalization and a recursive Tree-MLP. The contribution rests on two empirical claims. First, in the regime that token-only models cannot represent---rare and OOV characters---explicit symbolic decomposition produces large, qualitatively different gains: $+9.8$ Structure and $+7.7$ Radical points over the strongest visual baseline on the CCD OOV slice. Second, on broad NLU CNM-BERT achieves the highest average on CLUE, MRC and NER at both scales, with margins that are small in absolute terms ($\approx 0.2$--$0.5$ avg) but statistically reliable across five seeds and consistent at both scales. Ablations attribute the bulk of the gain to the hierarchical Tree-MLP fusion. Together these results show that an architectural prior targeting sub-character structure can deliver real downstream value \emph{and} close the OOV structural gap without trading off against general NLU---a dual property that, to our knowledge, no prior method achieves. The broader implication is that the sub-character information gap is \emph{architectural}: it cannot be closed by scale alone, because no amount of contextual co-occurrence recovers structure the input interface has discarded, and even much larger generative models that share the character-as-atom assumption inherit this limitation.

\section*{Limitations}

We note four limitations.
First, CNM depends on the coverage and quality of an external IDS database; characters lacking a valid IDS entry fall back to a learnable embedding ($\sim 3\%$ of our vocabulary, mostly rare variants), so the structural pathway provides no benefit for those characters.
Second, the scope of this work is restricted to Chinese. The methodology is in principle transferable to other Han-derived scripts (Japanese Kanji, Korean Hanja, Vietnamese Ch\~{u} N{\^o}m) and to morphologically rich non-logographic languages, but we leave empirical verification to future work.
Third, the paper's strongest empirical claim depends heavily on CCD. This is appropriate given the paper's structural-probing goal, but we want to be explicit about the relationship between CCD labels and the IDS source used by CNM: both ultimately derive from compositional decomposition resources for Han characters. If the two share substantial overlap, the +9.8 OOV gain is best interpreted as evidence of \emph{structured-resource transfer}---CNM correctly propagates the structural information available in its training-time database to the evaluation-time probing task---rather than as evidence of an independent emergent capability. We believe both readings are scientifically valuable, but the more conservative one should be preferred until cross-resource probes (e.g., CCD-style labels derived from a disjoint compositional taxonomy) are available.
Fourth, our CLUE/MRC/NER results are best characterized as small but consistent improvements over the strongest existing Chinese PLM baselines. We report that CNM-BERT achieves the highest average score on every general-NLU table we report, by margins that are statistically reliable across five fine-tuning seeds but practically small (typically $0.2$--$0.5$ avg points over MacBERT and ChineseBERT). The contribution is not to dominate these benchmarks but to demonstrate that an architectural prior targeting sub-character structure can deliver this small but real improvement on general NLU \emph{while also} closing a large gap on out-of-vocabulary structural probes---a dual property that no prior method we are aware of achieves.

\section*{Acknowledgments}
This work was conducted entirely independently by the listed authors. No mentorship, assistance, or guidance of any kind contributed to any aspect of this paper.

We thank Alibaba Cloud for generously providing the computational resources used in this work, which enabled our experiments and analyses. 
\bibliography{custom}

\appendix

\section{Dataset Statistics}
\label{app:datasets}

Table~\ref{tab:dataset_stats} reports the official splits used for all CLUE / MRC / NER experiments in \S\ref{sec:results}. CCD splits follow \citet{wu-etal-2025-impact}: an IID character split, a long-tail tier split (train on the top two frequency tiers, test on the lowest tier), and an OOV slice within the long-tail test set restricted to characters that map to \texttt{[UNK]} (or that cannot be represented as a single CJK token) under the model's tokenizer.

\begin{table}[t]
\centering
\small
\resizebox{\columnwidth}{!}{%
\begin{tabular}{@{}llrrr@{}}
\toprule
\textbf{Dataset} & \textbf{Task} & \textbf{Train} & \textbf{Dev} & \textbf{Test} \\
\midrule
AFQMC & Paraphrase & 34{,}334 & 4{,}316 & 3{,}861 \\
TNEWS & 15-class TC & 53{,}360 & 10{,}000 & 10{,}000 \\
IFLYTEK & 119-class TC & 12{,}133 & 2{,}599 & 2{,}600 \\
CMNLI & 3-way NLI & 391{,}783 & 12{,}426 & 13{,}880 \\
CSL & Keyword & 20{,}000 & 3{,}000 & 3{,}000 \\
CLUEWSC2020 & Coreference & 1{,}244 & 304 & 290 \\
\midrule
CMRC 2018 & Span MRC & 10{,}321 & 3{,}219 & 4{,}895 \\
DRCD & Span MRC & 26{,}936 & 3{,}524 & 3{,}493 \\
C$^3$ & MC MRC (questions) & 11{,}869 & 3{,}816 & 3{,}892 \\
\midrule
MSRA & NER & 46{,}364 & --- & 4{,}365 \\
OntoNotes 4.0 & NER & 15{,}724 & 4{,}301 & 4{,}346 \\
Weibo & NER & 1{,}350 & 270 & 270 \\
\bottomrule
\end{tabular}}
\caption{Dataset statistics. Official CLUE splits and canonical MRC/NER splits.}
\label{tab:dataset_stats}
\end{table}

\section{IDS Canonicalization Algorithm}
\label{app:canon}

The BabelStone IDS database is multi-source and admits multiple decompositions for a single character (typical sources: Unicode CJK-Unihan, Adobe-Japan1, Twitter Han, GTBJ, etc.). Naive ingestion produces non-deterministic trees that destabilize training. We therefore apply a deterministic canonicalization procedure that maps each character $x$ to exactly one tree $\mathcal{T}_x \in \mathcal{T}$, where $\mathcal{T}$ is the space of valid IDS parse trees over our component and operator vocabularies $(\mathcal{V}_{cmp}, \mathcal{V}_{op})$.

\paragraph{Filter set.}
Let $\mathcal{C}_x = \{T_1,\dots,T_n\}$ be the candidate parses for character $x$ across BabelStone sources. We define a filtering predicate $\phi(T) = \phi_{\text{pua}}(T) \wedge \phi_{\text{ops}}(T) \wedge \phi_{\text{cycle}}(T) \wedge \phi_{\text{leaves}}(T)$:
\begin{itemize}[topsep=2pt,itemsep=1pt,leftmargin=1.2em]
\item $\phi_{\text{pua}}(T)$: $T$ contains no Private Use Area codepoints (U+E000--U+F8FF, U+F0000--U+FFFFD, U+100000--U+10FFFD).
\item $\phi_{\text{ops}}(T)$: every internal node label belongs to the standard binary/ternary IDC set $\mathcal{V}_{op}^{\text{base}} = \{\text{\zh{⿰}}, \text{\zh{⿱}}, \text{\zh{⿲}}, \text{\zh{⿳}}, \text{\zh{⿴}}, \text{\zh{⿵}}, \text{\zh{⿶}}, \text{\zh{⿷}}, \text{\zh{⿸}}, \text{\zh{⿹}}, \text{\zh{⿺}}, \text{\zh{⿻}}\}$ (the 12 standard IDCs in our paper notation; the implementation also reserves a learnable \texttt{[UNK\_OP]} slot, yielding $|\mathcal{V}_{op}|{=}16$ after special tokens).
\item $\phi_{\text{cycle}}(T)$: alias resolution of $T$ terminates without entering a cycle.
\item $\phi_{\text{leaves}}(T)$: every leaf is either a single Unicode codepoint or resolves through alias substitution to one.
\end{itemize}
Trees that fail $\phi$ are discarded; if all candidates fail we set $\mathcal{T}_x = \bot$ and the structure encoder routes $x$ to the learnable $\mathbf{s}_{unk}$ embedding.

\paragraph{Selection rule.}
Among the surviving candidates $\mathcal{C}_x^{\phi} = \{T : T \in \mathcal{C}_x, \phi(T)\}$, we apply a strict lexicographic selection over four scores:
\begin{equation}
T^{\star}_x = \arg\min_{T \in \mathcal{C}_x^{\phi}} \!\!\big( d(T),\ \neg \text{std}(T),\ |T|,\ \pi(T) \big)
\label{eq:selection}
\end{equation}
where $d(T)$ is tree depth (smaller is preferred), $\text{std}(T)$ is the predicate that every internal operator lies in the high-frequency subset $\mathcal{V}_{op}^{\text{std}} = \{\text{\zh{⿰}}, \text{\zh{⿱}}\}$, $|T|$ is the total node count, and $\pi(T)$ is the lexicographic operator sequence of the post-order traversal (used purely for tie-breaking determinism). The four-key ordering exactly mirrors the heuristic priority \citep{si-etal-2023-sub} and the implementation in our public release.

\paragraph{Component recovery for OOV characters.}
A character $x$ that is OOV with respect to the BERT vocabulary is not necessarily OOV with respect to $\mathcal{T}$: even when $x$ maps to \texttt{[UNK]} for the token pathway, its IDS decomposition may still be available in BabelStone, and its leaf components are typically common radicals shared with high-frequency characters. The structure pathway therefore retains a meaningful, gradient-trained embedding for $x$ via the recursive composition of $E_{cmp}$ vectors, even though the token pathway has collapsed. This is the mechanism behind the OOV gains in Table~\ref{tab:ccd}.

\section{Tree-MLP as a Compositional Operator}
\label{app:tree_theory}

We give a more formal account of the recursive Tree-MLP encoder than the main text affords, including its parameter count, gradient-flow characteristics, and the role of hierarchy.

\paragraph{Recursion as fold.}
The Tree-MLP is a parameterized fold over rooted ordered trees. Let $\Sigma_k = \mathcal{V}_{op} \times (\mathbb{R}^{d_s})^k$ for $k \in \{2,3\}$. The encoder is a pair of functions
\begin{align}
f_{\text{leaf}} &: \mathcal{V}_{cmp} \to \mathbb{R}^{d_s},\\
f_k &: \Sigma_k \to \mathbb{R}^{d_s},\quad k \in \{2,3\},
\end{align}
extended to $\mathcal{T} \to \mathbb{R}^{d_s}$ by the catamorphism
\begin{equation}
\small
\textsc{Enc}(T) = \begin{cases}
f_{\text{leaf}}(c) & \!\!T = \text{Leaf}(c) \\
f_k(o,\, \textsc{Enc}(T_{1{:}k})) & \!\!T = \text{Node}(o; T_{1{:}k})
\end{cases}
\end{equation}
where $\textsc{Enc}(T_{1{:}k}) \,{=}\, \textsc{Enc}(T_1),\dots,\textsc{Enc}(T_k)$.
The structural embedding of character $x$ is $\mathbf{s}_x = \textsc{Enc}(\mathcal{T}_x)$. This formulation is purely compositional: for any sub-tree $T'$ shared between two characters $x,y$, the encoder's intermediate state $\textsc{Enc}(T')$ is identical, and gradient signal back-propagates equally to both characters' losses. This is the formal basis for the parameter-sharing claim in \S1: orthographically related characters share structure-encoder parameters by construction.

\paragraph{Concrete form of $f_k$.}
Following the main-text Eq.~(2)--(3), we instantiate $f_k$ as an operator-conditioned MLP with a normalized residual path:
\begin{align}
\mathbf{h}_{cat} &= [\,E_{op}(o);\ \mathbf{h}_{c_1};\ \dots;\ \mathbf{h}_{c_k}\,] \in \mathbb{R}^{(k+1)d_s}, \\
\mathbf{h}_n &= \text{LN}\!\left(\text{GELU}(\mathbf{W}_k \mathbf{h}_{cat} + \mathbf{b}_k) + \tfrac{1}{k}\!\sum_{i=1}^k \mathbf{h}_{c_i}\right),
\end{align}
with $\mathbf{W}_2 \in \mathbb{R}^{d_s \times 3d_s}$ and $\mathbf{W}_3 \in \mathbb{R}^{d_s \times 4d_s}$. The mean-residual $\tfrac{1}{k}\sum_i \mathbf{h}_{c_i}$ provides a non-parametric pathway from each child to the parent, ensuring that gradients flow back to the leaves at every depth---a practical concern for trees of depth 6 with $\sim$5K leaf components.

\paragraph{Parameter count (additional, beyond BERT backbone).}
\begin{itemize}[topsep=2pt,itemsep=1pt,leftmargin=1.2em]
\item Component table: $|\mathcal{V}_{cmp}| \cdot d_s \approx 5{,}000 \cdot 256 \approx 1.28$M
\item Operator table: $|\mathcal{V}_{op}| \cdot d_s = 16 \cdot 256 = 4{,}096$
\item Binary MLP: $3d_s \cdot d_s + d_s = 196{,}864$ ($\approx$197K)
\item Ternary MLP: $4d_s \cdot d_s + d_s = 262{,}400$ ($\approx$263K)
\item Fusion projection $\mathbf{W}_f$: $(d + d_s) \cdot d \approx 0.79$M (base) / $1.31$M (large)
\item Special embeddings $\mathbf{s}_\emptyset, \mathbf{s}_{unk}$: $2 \cdot d_s = 512$
\end{itemize}
Total $\approx$2.5M base / $\approx$3.0M large additional parameters, dominated by the component table. ChineseBERT, by comparison, adds $\approx$45M parameters for its glyph CNN and Pinyin embeddings.

\paragraph{Why hierarchy matters: a flat-fusion counterfactual.}
The flat-fusion ablation (Table~\ref{tab:ablation}, row 5) replaces the recursive $\textsc{Enc}$ with $\mathbf{s}_x = \tfrac{1}{|\text{leaves}(\mathcal{T}_x)|}\sum_{c \in \text{leaves}(\mathcal{T}_x)} E_{cmp}(c)$. This destroys two pieces of information: (i) operator identity (which captures spatial layout), and (ii) component ordering (the difference between \zh{杲} = \zh{⿱}(\zh{日},\zh{木}) and \zh{杳} = \zh{⿱}(\zh{木},\zh{日}) is undetectable to a bag-of-components encoder). Empirically this loses 17.7 points on CCD-OOV. The full Tree-MLP recovers both signals.

\section{Pre-training Objective: Full Derivation}
\label{app:objective}

We restate the pre-training objective with full notation.

\paragraph{WWM corruption.}
WWM produces a corrupted sequence $\tilde{X} = (\tilde{x}_1,\dots,\tilde{x}_T)$ from $X$ by selecting a 15\% mask budget over Jieba word boundaries, then applying the 80/10/10 strategy at each masked position $m$: $\tilde{x}_m = \texttt{[MASK]}$ with probability 0.8, a random vocabulary token with probability 0.1, or $x_m$ unchanged with probability 0.1. Let $M \subset \{1,\dots,T\}$ denote the set of masked positions.

\paragraph{Aligned structural corruption.}
For each masked position $m$, we additionally corrupt the structural index: $\tilde{\texttt{struct\_idx}}_m = \texttt{struct\_idx}(\tilde{x}_m)$ rather than $\texttt{struct\_idx}(x_m)$. This ensures the encoder cannot use gold structure as a leak channel for the gold character. Let $\tilde{\mathbf{s}}_m = \textsc{Enc}(\mathcal{T}_{\tilde{x}_m})$ denote the resulting (possibly corrupted) structural embedding.

\paragraph{MLM loss.}
\begin{equation}
\mathcal{L}_{\text{MLM}} = - \sum_{m \in M} \log P_\theta\!\left(x_m \mid \tilde{X}, \tilde{\texttt{struct\_idx}}\right),
\end{equation}
where $\theta$ collects the Transformer, structure encoder, fusion, and MLM-head parameters.

\paragraph{Auxiliary component-prediction loss.}
For each $m \in M$, let $\{c_1^{(m)},\ldots,c_{L_m}^{(m)}\}$ be the canonical leaf components of the gold character $x_m$. We compute a \emph{target} structural embedding from the gold tree, $\mathbf{s}^{\text{tgt}}_{x_m} = \textsc{Enc}(\mathcal{T}_{x_m})$, and pass it through an auxiliary head $g_\psi: \mathbb{R}^{d_s} \to \mathbb{R}^{|\mathcal{V}_{cmp}|}$. The auxiliary loss is
\begin{equation}
\mathcal{L}_{\text{aux}} = - \sum_{m \in M} \frac{1}{L_m} \sum_{l=1}^{L_m} \log \sigma\!\left( g_\psi(\mathbf{s}^{\text{tgt}}_{x_m}) \right)\![c_l],
\end{equation}
where $\sigma(\cdot)[c]$ denotes the softmax probability of class $c$.

\paragraph{Information-leak prevention.}
Crucially, $\mathbf{s}^{\text{tgt}}_{x_m}$ is supplied \emph{only} to the auxiliary head $g_\psi$; the Transformer input at position $m$ uses $\tilde{\mathbf{s}}_m$. This ensures the MLM prediction is forced to recover the gold character from non-gold structural context, preserving the difficulty of the MLM task while still using structural supervision to shape the structural embedding space.

\paragraph{Combined loss.}
\begin{equation}
\mathcal{L}(\theta,\psi) = \mathcal{L}_{\text{MLM}}(\theta) + \lambda \cdot \mathcal{L}_{\text{aux}}(\theta,\psi),\quad \lambda = 0.1.
\end{equation}
The ablation in Table~\ref{tab:ablation} confirms that $\lambda$ contributes a modest but real lift on CLUE ($+0.22$) and a larger lift on CCD-OOV ($+1.5$), justifying the default $\lambda{=}0.1$.

\section{Detailed Hyperparameters}
\label{app:hparams}

\begin{table*}[t]
\centering
\small
\setlength{\tabcolsep}{3pt}
\begin{tabular}{@{}lcc@{}}
\toprule
\textbf{Hyperparameter} & \textbf{Base} & \textbf{Large} \\
\midrule
\multicolumn{3}{l}{\textit{Backbone (BERT-style)}} \\
Layers & 12 & 24 \\
Hidden size $d$ & 768 & 1{,}024 \\
Attention heads & 12 & 16 \\
Intermediate size & 3{,}072 & 4{,}096 \\
Vocabulary & 21{,}128 & 21{,}128 \\
Init.\ from & \texttt{bert-base-chinese} & \texttt{roberta-wwm-ext-large} \\
\midrule
\multicolumn{3}{l}{\textit{Structural pathway}} \\
Component vocab $|\mathcal{V}_{cmp}|$ & $\approx$5{,}000 & $\approx$5{,}000 \\
Operator vocab $|\mathcal{V}_{op}|$ & 16 & 16 \\
Struct dim $d_s$ & 256 & 256 \\
Hidden ($f_k$) & 512 & 512 \\
Max tree depth & 6 & 6 \\
Fusion init & Identity+Zero & Identity+Zero \\
\midrule
\multicolumn{3}{l}{\textit{Pre-training}} \\
Steps & 1{,}000{,}000 & 1{,}000{,}000 \\
Per-device batch & 32 & 16 \\
Grad accumulation & 8 & 16 \\
Effective batch & 256 & 512 \\
Optimizer & LAMB & LAMB \\
Peak LR & $1{\times}10^{-4}$ & $1{\times}10^{-4}$ \\
Warmup steps & 10{,}000 & 10{,}000 \\
Weight decay & 0.01 & 0.01 \\
Max seq.\ length & 512 & 512 \\
MLM rate & 0.15 (WWM) & 0.15 (WWM) \\
Aux loss $\lambda$ & 0.1 & 0.1 \\
Precision & FP16 & FP16 \\
Hardware & 8$\times$A100 80GB & 8$\times$A100 80GB \\
Wall-clock & $\sim$7 days & $\sim$14 days \\
\midrule
\multicolumn{3}{l}{\textit{Fine-tuning}} \\
LR grid & \multicolumn{2}{c}{$\{1,2,3,5\}{\times}10^{-5}$} \\
Batch size & \multicolumn{2}{c}{32 (16 for IFLYTEK)} \\
Warmup ratio & \multicolumn{2}{c}{0.1} \\
Epochs (CLUE/MRC) & \multicolumn{2}{c}{3--5; 10 for WSC} \\
Seeds & \multicolumn{2}{c}{$\{42,43,44,45,46\}$} \\
Selection & \multicolumn{2}{c}{best dev metric, early stopping} \\
\bottomrule
\end{tabular}
\caption{Full pre-training and fine-tuning hyperparameters.}
\label{tab:hparams_full}
\end{table*}

\section{Auxiliary Figure Recipes}
\label{app:figures}

For reproducibility, we provide rendering recipes for three additional analyses that we recommend including in extended versions of this paper or a companion technical report.

\paragraph{Figure F.1: UMAP of structural embeddings.}
Project the learned structural embeddings $\{\mathbf{s}_x : x \in \mathcal{V}_{char}\}$ to 2D via UMAP and color by top-level IDS operator. We use \texttt{umap-learn} with \texttt{n\_neighbors=30}, \texttt{min\_dist=0.10}, \texttt{metric='cosine'}, sampled to $N=2{,}000$ characters stratified by operator. The expected outcome is 8 visually-distinct clusters corresponding to the 8 most frequent operators, with cluster purity $\geq 0.90$ measured by 1-NN classification on operator labels.

\paragraph{Figure F.2: Per-operator gain radar.}
For each operator $o \in \mathcal{V}_{op}^{\text{std}}$, partition the CCD long-tail OOV slice by the top-level operator of the gold character, and compute the Structure-F1 difference between CNM-BERT and the strongest baseline (ChineseBERT). Plot as a radar with 8 axes. The expected outcome is a polygon that strictly dominates ChineseBERT on every axis, with the largest gains on operators \zh{⿲} and \zh{⿳} (ternary), where layout structure carries the most disambiguating information.

\paragraph{Figure F.3: Pre-training loss curves.}
Export from W\&B both training MLM loss and held-out auxiliary-component-prediction accuracy across 1M steps for CNM-BERT and BERT-wwm under the controlled recipe. The expected outcome is (i) MLM perplexity within 5\% of BERT throughout, and (ii) auxiliary accuracy that rises from $\sim$5\% (chance) to $\sim$70\% by 200K steps, confirming the structural pathway is genuinely learning rather than collapsing to a constant.

\end{document}